\documentclass[11pt]{amsart}

\usepackage{graphicx,float,algpseudocode,array,amsmath,amssymb}
\usepackage[caption = false]{subfig}

\graphicspath{ {./ims/} }

\begin{document}




\title{Generative Simultaneous Localization and Mapping (G-SLAM)}


\author{Nikos Zikos}
\address{Department of Electrical and Computer Engineering, Aristotle University of Thessaloniki, 
Greece}
\email{nzikos@auth.gr}

\author{Vassilios Petridis}
\email{petridis@eng.auth.gr}

\begin{abstract}
Environment perception is a crucial ability for robot's interaction into an environment. One of the first steps in this direction is the combined problem of simultaneous localization and mapping (SLAM). A new method, called G-SLAM, is proposed, where the map is considered as a set of scattered points in the continuous space followed by a probability that states the existence of an obstacle in the subsequent point in space. A probabilistic approach with particle filters for the robot's pose estimation and an adaptive recursive algorithm for the map's probability distribution estimation is presented. Key feature of the G-SLAM method is the adaptive repositioning of the scattered points and their convergence around obstacles. In this paper the goal is to estimate the best robot trajectory along with the probability distribution of the obstacles in space. For experimental purposes a four wheel rear drive car kinematic model is used and results derived from real case scenarios are discussed.

\end{abstract}





\maketitle
\section{Introduction}

The problem of Simultaneous Localization And Mapping (SLAM) is vital in case of  autonomous robots and vehicles navigating in unknown environments \cite{57472}. Usually the map consists of a sequence of features (or landmarks), each one of which represents the position of an obstacle or a part of an obstacle (i.e. a big obstacle can be represented by many features). 

The Extended Kalman Filter (EKF) was extensively used in the SLAM problem \cite{Smith:1986fk}, but it has the disadvantage that the computational cost increases significantly with the number of features. 
Since then, many probabilistic approaches have proposed \cite{Thrun:2005fk} including the Montemerlo's et al. solution to stochastic SLAM, FastSLAM 1.0 and 2.0 \cite{1241885,DBLP:conf/aaai/MontemerloTKW02,DBLP:conf/ijcai/MontemerloTKW03,Montemerlo:2007kx},  Grid-based SLAM \cite{Dissanayake:2001fk,Grisetti:2007vn}, Dual-FastSLAM \cite{Rodriguez-Losada:fk}, DP-SLAM \cite{dpslam}, L-SLAM \cite{Zikos:2014aa,mywcci10}, etc.

In mobile robots 2-D maps are often sufficient, especially when a robot navigates on a flat surface and the sensors are mounted so that they capture only a slice of the world.

Instead of occupancy grid maps with fine-grained grid defined over the continuous space, in this paper a set of scattered points in the continuous space is used. 
It is presented a new method, called G-SLAM \cite{6265692}, where the map is considered as a set of scattered points in the continuous space followed by a probability that states the existence of an obstacle in the subsequent point in space. In addition to \cite{6265692} we have previously presented, in this paper it is presented the mathematical formulation and derivation of the problem and the experiments and the results are enhanced with more state of the art SLAM methods.

A probabilistic approach based on particle filters is used for the robot's pose estimation. In the robot's pose estimation the time series of controls, the time series of the measurements and the latest estimation of the probabilistic map are involved. 

A recursive algorithm for the map's probability distribution estimation is used for the map update procedure.
The proposed method generates new hypothetical points of features in space which are subsequently tested whether they correspond to real obstacles or not. That is why we call it G-SLAM for Generative-SLAM.

Key feature of the G-SLAM method is the adaptive repositioning of the scattered map's points that results in a convergence of all the points around obstacles. The final map resulted from the G-SLAM method exhibits high density of weighted points around the obstacle and a subsequent high sparsity in the space which is free of obstacles. These weighted points represent the probability distribution of the obstacles in the continuous space. This method fits on problems where a detailed map is needed with low computational resources.

This paper is organized as follows. In section \ref{s3} the probabilistic analysis of the combined SLAM problem in terms of recursively computed probability distributions which estimates the probabilistic map and the robot's trajectory along with the G-SLAM method are presented. In section \ref{s2} the model of the robotic system which consist of the robot's kinematic model and the distance-bearing measurement model is described. In section \ref{sResults} experimental results from real case scenario are discussed.


\section{SLAM problem definition}
\label{s3}

\subsection{Notations}

\begin{itemize}
	\item $s^t$: is a time series of the robot's pose, while $s_t$ is only the pose at a time instance.
	\item $\Theta$: represents the map and is a set of points $\theta^k$ in space.
	\item $z^t$: is a time series of the measurements, while $z_t$ represents only the measure at time $t$.
	\item $u^t$ is the time series of the robot's control inputs, while $u_t$ refers only at time $t$ control input.
	\item $f(.)$: is the robot's kinematic model. 
	\item $g(.)$: represents the sensor's measurement model.
	\item $d_z(.)$: represents the probability density function of the sensor's measure noise.
	\item $d_f(.)$: represents the probability density function of the transition's model noise.
\end{itemize}

\subsection{SLAM Posterior}

While most SLAM methods are trying to estimate the robot's pose $s_t$ and the map $\Theta$ at timestamp $t$, in this paper our goal is to estimate the whole time series $s^t$ and the map $\Theta$ using the observation time series $z^t$ and the control time series $u^t$. In probabilistic terms this  posterior is expressed as:

\begin{equation}
\label{eq.SLAM}
Prob(s^{t},\Theta | u^{t},z^{t})
\end{equation}

Using the definition of the conditional probability, the posterior in equation \ref{eq.SLAM} can be expressed as:

\begin{equation}
\label{eq.factored}
Prob(s^{t},\Theta | u^{t},z^{t})=
\underbrace{Prob(s^{t}|z^{t},u^{t})}_\text{trajectory posterior}
\underbrace{Prob(\Theta|s^{t},z^{t},u^{t})}_\text{map posterior}
\end{equation}

The two factors of the equation \ref{eq.factored} correspond to the robot's trajectory posterior and the map's posterior respectively. The calculation of these two factors is discussed bellow.


\subsection{Pose prediction}
\label{s.posepr}

The left posterior of the equation \ref{eq.factored} refers to the estimation of the pose time series given the map and the time series of the observation and the controls.

The calculation of this posterior is done using the technique of particle filtering. The proposal distribution for the particles will be the posterior $p(s^{t}|z^{t-1},u^{t})$, thus the drawing process for each particle $i$ evolves only the previous state $s_{t-1}^i$ and the current control input $u_{t}$.

\begin{equation}
\label{eq.pose}
s_{t}^i \sim p(s_{t}|s_{t-1}^i,u_{t})
\end{equation}

The proposal distribution is generated from the posterior $Prob(s_{t}|s_{t-1},u_{t})$ using the robot's kinematic model $f$ and of course a random sample of the control's input noise $\epsilon_{t}^i$.

\begin{equation}
s_{t}^i = f(s_{t-1}^i,u_{t},\epsilon_{t}^i)
\end{equation}

This procedure creates a cloud of $N$ particles, all representing a possible pose of the robot. It is noteworthy that each particle carries out its own estimation of the map which is independent from the other particles' maps. The estimation of the pose timeseries is not done yet since particle filter's importance factors have to be calculated. The calculation of the importance factors is discussed below in the section \ref{s.importancef}.

\subsection{Map Update}
\label{s.mapupdate}

The rightmost factor of the equation \ref{eq.factored} refers to the estimation of the map given the time series of the observation and the controls. The map consists of a set of scattered points in space $\theta$ and each one is associated with a probability that the point $\theta$ is an obstacle. The distribution of this probabilistic map can be represented by the following posterior.

\begin{equation}
\label{eq.mappost}
p_{t}^k=Prob(\theta_k | s^{t},u^{t},z^{t})
\end{equation}

The equation \ref{eq.mappost} gives the probability that the feature $\theta_k$ is an obstacle given the observations $z^{t}$ and the controls $u^{t}$. By the definition of the conditional probability, the posterior \ref{eq.mappost} is expressed as:

\begin{equation}
\label{eq.mup2}
p_{t}^k=\frac{Prob(z_{t},\theta_k | s^{t},u^{t},z^{t-1})}{Prob(z_{t} | s^{t},u^{t},z^{t-1})}
\end{equation}

Using the law of total probability for all $\theta_j$ the denominator becomes

\begin{equation}
\label{eq1}
p_{t}^k=\frac{Prob(z_{t},\theta_k | s^{t},u^{t},z^{t-1})}{\sum_j{Prob(z_{t} ,\theta_j| s^{t},u^{t},z^{t-1})}}
\end{equation}

The posteriors of the numerator and the denominator have the same form

\begin{align*}
&Prob(z_{t},\theta_k | s^{t},u^{t},z^{t-1})=\\
&Prob(z_{t} | s^{t},u^{t},z^{t-1},\theta_k)Prob(\theta_k | s^{t},u^{t},z^{t-1})
\end{align*}

In the rightmost posterior we note that $\theta_k$ is independent of the control input $u_{t}$ and the pose $s_{t}$ due to the absence of the measurement $z_{t}$. Thus the above expression becomes

\begin{align}
\label{eq3}
&Prob(z_{t} | s^{t},u^{t},z^{t-1},\theta_k)Prob(\theta_k | s^{t-1},u^{t-1},z^{t-1})=\nonumber\\
&Prob(z_{t} | s^{t},u^{t},z^{t-1},\theta_k)p_{t-1}^k
\end{align}

Equations \eqref{eq1} and \eqref{eq3} imply the recursion:

\begin{equation}
\label{eq4}
p_{t}^k=\frac{Prob(z_{t}| s^{t},u^{t},z^{t-1},\theta_k)p_{t-1}^k} {\sum_j{Prob(z_{t}| s^{t},u^{t},z^{t-1},\theta_j)p_{t-1}^j}}
\end{equation}

Since $z_t$ is independent from previous observations, control inputs and previous robot's positions the equation \eqref{eq4} is simplified as:

\begin{equation}
\label{eq5}
p_{t}^k=\frac{Prob(z_{t}| s_{t},\theta_k)p_{t-1}^k} {\sum_j{Prob(z_{t}| s_{t},\theta_j)p_{t-1}^j}}
\end{equation}

In order to compute the recursion \eqref{eq5} we need to calculate the quantity $q^k_{t}=Prob(z_{t}| s_{t},\theta_k)$. In case that the probability distribution function of the measurement noise is given by function $d_z(z)$, then this quantity can be calculated by:

\begin{equation}
\label{eq6}
q^k_{t} = d_z \left( g(s_{t},\theta_k) \right)
\end{equation}

Combining equations \eqref{eq5} and \eqref{eq6} the probability of every point $k$ is calculated using equation \eqref{eq7}.

\begin{equation}
\label{eq7}
p_{t}^k=\frac{q^k_{t}p_{t-1}^k} {\sum_j{q^j_{t}p_{t-1}^j}}
\end{equation}

\subsection{Importance factor calculation}
\label{s.importancef}

The distribution as proposed in equation \eqref{eq.pose} is only the proposal distribution. Using the simulation technique of particle filtering the target distribution is calculated as:

\begin{equation*}
\text{target distribution} = \text{proposal distribution} * \text{importance factor}
\end{equation*}

Through the target distribution the best estimation for the robot's pose $s_{t}$ is calculated.

\begin{equation}
\label{wfactor1}
w_{t}^i=\frac{\text{target distribution}}{\text{proposal distribution}}=
\frac{p(s^{t,i}|z^{t},u^{t})}{p(s^{t,i}|z^{t-1},u^{t})}
\end{equation}

Using the Bayes Theorem the equation \eqref{wfactor1} is simplified as:

\begin{equation}
w_{t}^i \propto \frac{p(s^{t,i}|z^{t-1},u^{t}) p(z_{t}|s^{t,i},u^{t},z^{t-1})}
{p(s^{t,i}|z^{t-1},u^{t})}
\end{equation}

\begin{equation}
w_{t}^i \propto p(z_{t}|s^{t,i},u^{t},z^{t-1})
\end{equation}

So the importance factor is proportional to the posterior $p(z_{t}|s^{t,i},u^{t},z^{t-1})$ which is already calculated in the map update section \ref{s.mapupdate} as the denominator of equations \eqref{eq.mup2} or \eqref{eq5}.

\begin{equation}
w_{t}^i \propto \sum_j{Prob(z_{t}| s_{t,i},\theta_{j,i})p_{t-1}^{j,i}}
\end{equation}

Since the set of particles $S^{t}=\{s^t_1,...,s^t_M\}$ is finite, the "cloud" of particles is growing as the time increases, which can lead to the degeneracy of the algorithm. Thus a resampling technique is necessary. In this paper and on the experiments that took place, the technique of Residual Systematic Resampling (RSR) is used \cite{1348894}.

\subsection{The G-SLAM Method}

The proposed method, G-SLAM is based on a technique that generates stochastically new scattered points that are added into the map. This stochastic generation is based on the current particle and the current observation. Then the update procedure updates the map by updating each point's probability, while afterwards the "meaningless" features are removed from the map set. The sensor's noise is responsible for the stochastic nature of this procedure. This addition of scattered point into the map set is achieved using a drawing procedure which is described bellow. Afterwards the extended map is updated and the updated points with low probabilities are removed. The small probability in a map point, states that this point in space is unlikely to be an obstacle. Algorithms of this type converges as are discussed in \cite{Kehagias91convergenceproperties}. In the context of this paper, map update procedure converges to high probabilities in map points near obstacles. Removing all low probability map points, the parts of space which are free of obstacles are also free of map points while on the other hand the parts of space with obstacles gathers all the scattered points around them.

The G-SLAM method can be described abstractly in six steps:

\begin{enumerate}
	\item Draw pose $s_{t}^i$ for every particle $i$ using the subsequent pose $s_{t-1}^i$ and the control $u_{t}$
	\item Generate and add new map points $\theta$ into the particle's $i$ map set  $\Theta^i$ using drawing process based on the observation $z_{t}$ and pose $s_{t}^i$
	\item Update map by calculating the probabilities of all map points
	\item Remove the map points with low probabilities
	\item Calculate Importance factors for every particle
	\item Resample particles if necessary
\end{enumerate}

Steps 1,3,5,6 are already discussed in sections \ref{s.posepr},\ref{s.mapupdate} and \ref{s.importancef}, while generation and removal of map points are discussed bellow.

\subsubsection{Updating existing map points}

The existing map can be easily updated using equation \eqref{eq7}. For every map point $\theta_i$ it is calculated the probability of the measure $z_t$ to correspond to this point in map using the equation \eqref{eq6} and then it is multiplied to the previous map's point probability $p_{t-1}^i$. It is noteworthy that the denominator of the equation \eqref{eq7} is just a normalization factor.

\subsubsection{Adding new map points}

The stochastic addition of new points into the map is achieved based on the observation $z_{t}$ and the current pose of each particle.
For every observation $z_{t}$ a drawing procedure takes place in order to generate a set of $M$ new map points that represents the sensor's measurement probability distribution.

\begin{equation}
\hat{z}_{t}^m \sim \mathcal{N} (z_{t};z_{t},R_{t})
\end{equation}
where $R_{t}$ is the covariance matrix of the sensor's noise.

Every element $\hat{z}_{t}^m$ is given a probability:
\begin{equation}
\hat{q}^m = d_z( \hat{z}_{t}^m)
\end{equation}

where $d_z(.)$ represents the probability density function of the measurement's noise.

These elements $\hat{z}^m_{t}$ are unlikely to correspond to a map point $\theta\in\Theta$. Thus in order to update the map it is necessary to create new map points $\hat{\theta}_m=g(s_t,\hat{z}^m_{t})$ in the map $\Theta$. But also, in order to proceed with the update, it is necessary to calculate each point's probability $\hat{p}^m_{t-1}$. In the G-SLAM method the calculation of the probability $\hat{p}^m_{t-1}$ is done numerically using interpolation methods. The pre-updated map contains points and their subsequent probabilities at time $t-1$. These points in space are interpolated with $\hat{\theta}_m$ in order to estimate the subsequent $\hat{p}^m_{t-1}$ probability. Afterwards and using the equation \eqref{eq7} the new map points are updated and added to the map set $\Theta$.
This procedure augments the probabilistic map with $M$ new points and their probabilities.

\subsubsection{Removing meaningless map points}

As already discussed, update procedure returns an augmented map with more map points than previously had. Some of those points might have near zero probability meaning that the probability of an obstacle existence in this point in space is highly unlikely.

In G-SLAM a map point removal procedure takes place after map update procedure in order to remove all the meaningless map points. So all points $\theta_i$ which their probabilities $p^i_t$ are less than a predefined thresshold $p^i_t<p_{thr}$ are removed from the map set. Using this technique it is prevented the overpopulation of the set $\Theta$ and the features $\theta$ tends to gather near obstacles. 


\subsection{Agorithm}

A pseudo code of the G-SLAM algorithm is given below. 

\

\begin{algorithmic}
	\Function{G-SLAM}{$s_{t-1},z_{t},u_{t},\Theta$}
	\For{$i=1:N$ particles}
	\State draw $s_{t}^i \sim f(s_{t-1}^i,u_{t})$ 
	\Comment{Drawing proccess}
	\For{k=1:all $\theta \in \Theta$}
	\Comment{Existing map update}
	\State $q^k_{t} = d_z \left( g(s_{t}^i,\theta_k) \right)$
	\State $p_{t}^k=\frac{q^k_{t}p_{t-1}^k} {\sum_j{q^j_{t}p_{t-1}^j}}$
	\EndFor
	\For{m=1:M}
	\Comment{Adding new map points}
	\State $\hat{z}_{t}^m \sim \mathcal{N} (z_{t};z_{t},R_{t})$
	\State $\hat{q}^m = d_z( \hat{z}_{t}^m)$
	\State $\hat{\theta}_m=g(s_t^i,\hat{z}^m_{t})$
	\State $p^m_{t-1}$ interpolate using $\Theta$ and $\hat{\theta}_m$
	\State $p_{t}^m=\frac{q^m_{t}p_{t-1}^m} {\sum_j{q^j_{t}p_{t-1}^j}}$
	\State Add $\hat{\theta}_m$ and $p_{t}^m$ in $\Theta$
	\EndFor
	\State $w_i=\sum_j{q^j_{t}p_{t-1}^j}$
	\Comment{Importance Factor Caclulation}
	\State Remove $\theta_k$ where $p_t^k<p_{thr}$
	\Comment{Meaninless point removal}
	\EndFor
	\State Resample particles
	\EndFunction
\end{algorithmic}


\section{Experiments \& Results}
\label{s4}
For experimental purposes the dataset performed by Nieto, Nebot et al. from the University of Sydney \cite{1241630,Nieto04thehybrid} was used. All the experiments were performed on this dataset with the car performing a full loop (fig. \ref{f1}). A four-wheel rear-drive car was used in this dataset. The car was equipped with a horizontal scanning laser sensor with 80 meters observing radius and 180 degrees field of view. The control vector of this car consists of the linear velocity of the rear left wheel and the heading angle of the front wheels. Also, GPS measurements comes with the dataset, which were used for the car's position validation.

\subsection{System description}
\label{s2}

\begin{figure}[h!]
	\centerline{\includegraphics[width=3.25in]{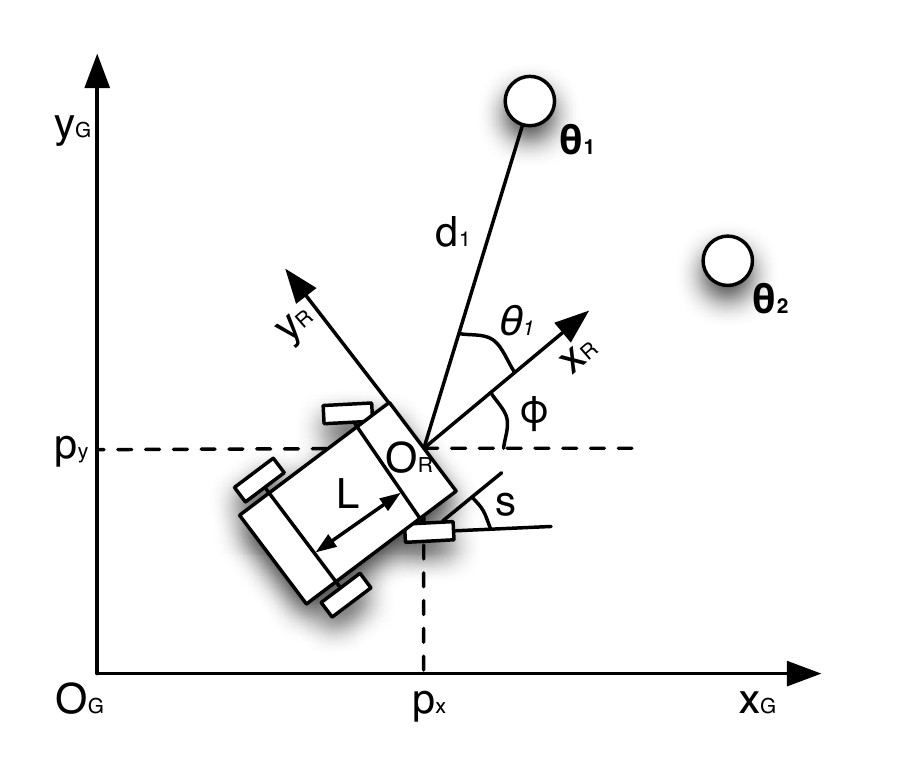}}
	\caption{The robot coordinate system and the position of $\theta_i$ landmark with respect to the robot coordinate system $O_R$. $\theta$ is the angle between the vectors $X_R$ and $O_R$\textbf{$\theta$}}
	\label{f.car}
\end{figure}

This dataset is a two dimensional planar dataset and the map is considered as a set of map point $\Theta=[\theta_1,\theta_2,....,\theta_N]$ each one corresponding to a point in space (fig. \ref{f.car}). The probability that a point $\theta_k$ is an obstacle is denoted by $p^k$. The set of features along with their probabilities is a probabilistic map of the space. The robot's path is represented by a time-series of it's pose $s^t=[s_1,s_2,...s_t]$.  In a planar problem each feature $\theta_k$ is a vector with entries $(x,y)$ coordinates of the point. Robot's pose $s_t$ is also a vector with entries $(x,y,\varphi)$ at time $t$ where $\varphi$ represents the angle of the robot's orientation corresponding to a global axis system. 

The measurement timeseries $z^t=[z_1,z_2,...z_t]$ consist of distance bearing measures $(d,\vartheta)$ acquired from the laser sensor and corresponds to the sensor's coordinate system. The distance-bearing laser sensor's feedback consists of a 361 distance measurements with half a degree angular distance between them.

\subsection{Model}

The car used for this experiments was a rear drive car-like four-wheel. The kinematic model of a vehicle like this is described by equation \eqref{eq.kinem}.

{\footnotesize 
	\begin{equation}
	\label{eq.kinem}
	\left[
	\begin{array}{c}
	p_{x,t+1}\\
	p_{y,t+1}\\
	\varphi_{t+1}
	\end{array}
	\right]=
	\left[
	\begin{array}{c}
	p_{x,t} + (v_c \cos(\varphi_{t}) - (a\sin(\varphi_{t})+b\cos(\varphi_{t}))\frac{v_c}{L}\tan(\omega))         \Delta t\\
	p_{y,t} + (v_c \sin (\varphi_{t}) + (a\cos(\varphi_{t})-b\sin(\varphi_{t}))\frac{v_c}{L}\tan(\omega))          \Delta t\\
	\varphi_{t} + \frac{v_t \Delta t}{L}\tan(\omega)
	\end{array}
	\right]
	\end{equation}}
where $v_c$ is the robot's linear velocity at time $t$, $\omega$ is the steering angle at time $t$, $L$ is the distance between the car's axles and $a,b$ are the coordinates of the laser sensor according to the car's coordinate system. The velocity $v_c$ is a function of the rear wheel's linear velocity and depends on the steering angle.

\begin{equation*}
v_c=\frac{v_e}{1-\tan(\omega)\frac{H}{L}}
\end{equation*}
where $H$ is the distance between the center point of the rear axle and the rear wheel.

The measurement model of a distance-bearing sensor is given by the nonlinear equations:

\begin{align}
z_t=&g(s_t,\theta_n)=\nonumber\\
=&\left[
\begin{array}{c}
\sqrt{(p_{x,t}-\vartheta_{x,n})^2+(p_{y,t}-\vartheta_{y,n})^2}\\
\arctan(\frac{p_{y,t}-\vartheta_{y,n}}{p_{x,t}-\vartheta_{x,n}})-\varphi_t
\end{array}
\right] \label{eq:meas1}
\end{align}

It is assumed that the distance-bearing sensor's measurements and the control measurements are noisy with noise functions of a known probability distributions.

\subsection{Results}
\label{sResults}
The algorithm was implemented using the kinematic model of equation \ref{eq.kinem}. The parameters that defines the car's kinematic model are: $L=2.75$, $H=0.74$, $a=L+0.5$ and $b=0.5$.

\begin{figure}[t]
	\centerline{
		\includegraphics[width=2.2in]{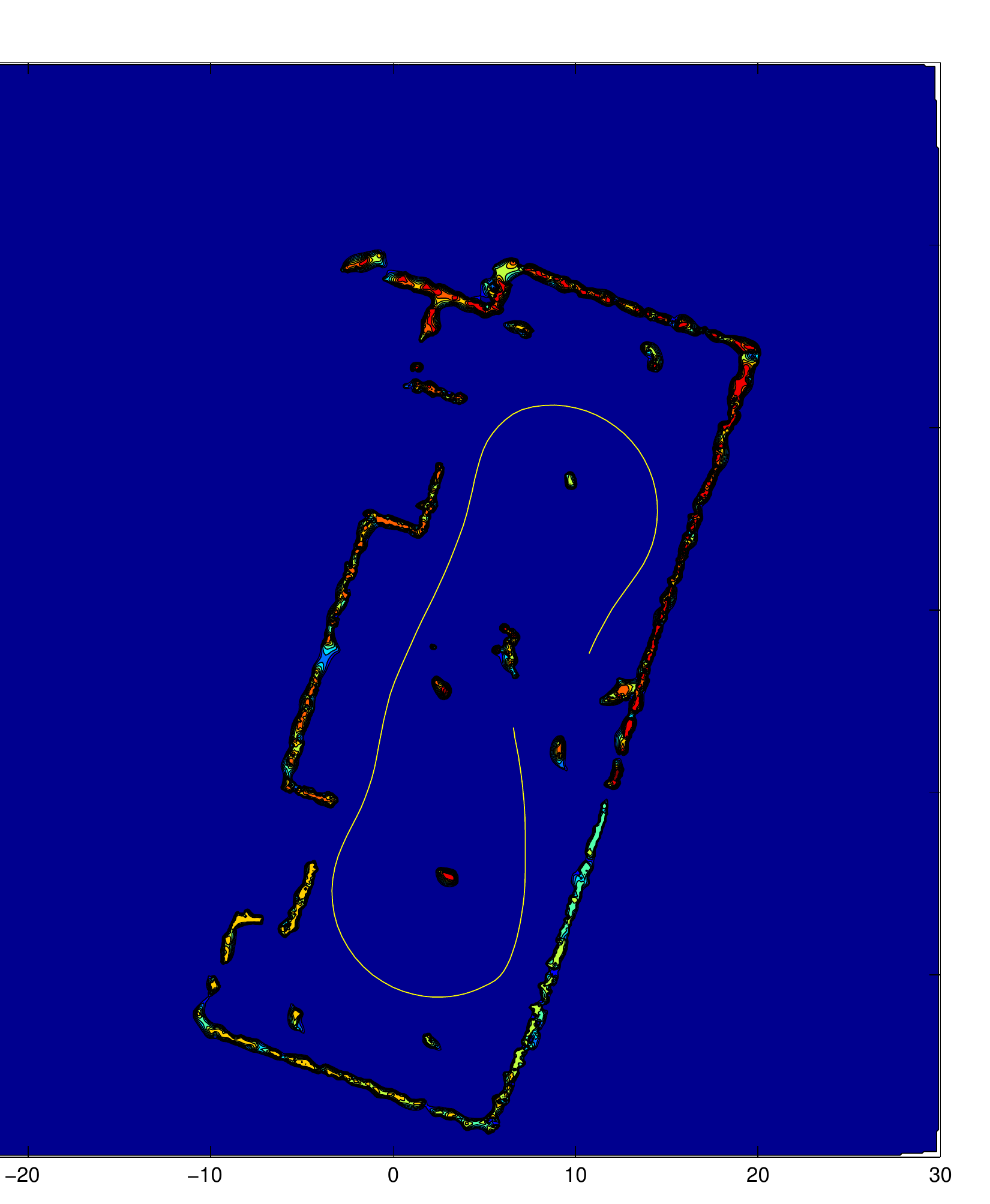}
		\includegraphics[width=2.2in]{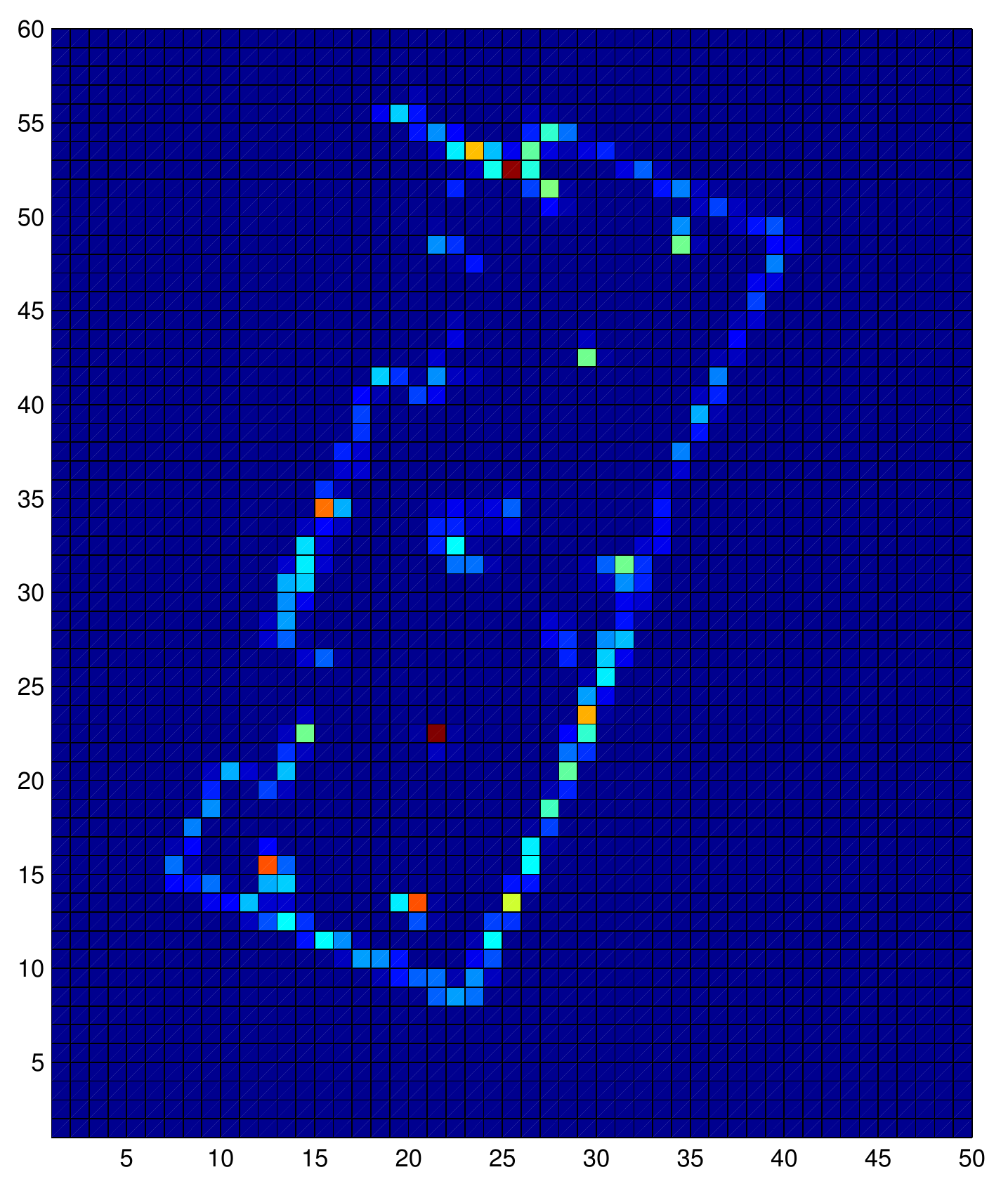}}
	\caption{a. Contour graph of the map's probability distribution using G-SLAM. The yellow line represents the car's path estimation. b. The map represented using Grid based SLAM. The resolution of the grid (3000 cells) was chosen to match the G-SLAM's population (3280 map points)}
	\label{f1}
\end{figure}

The algorithm results in a probabilistic map that consists of a set of points in space and their probabilities of being an obstacle.
Figure \ref{f1} shows a contour graph with the map's probability distribution in contrast with the Grid Occupancy SLAM with almost the same number of map points. The yellow line represents the best estimation for the car's path. G-SLAM method resulted 3280 map points all of them gathered around obstacles, while Grid occupancy SLAM with 3000 cells resulted a lower resolution map since most of the cells covers an area free of obstacles. 

\begin{figure}
	\centerline{
		\includegraphics[width=2.25in]{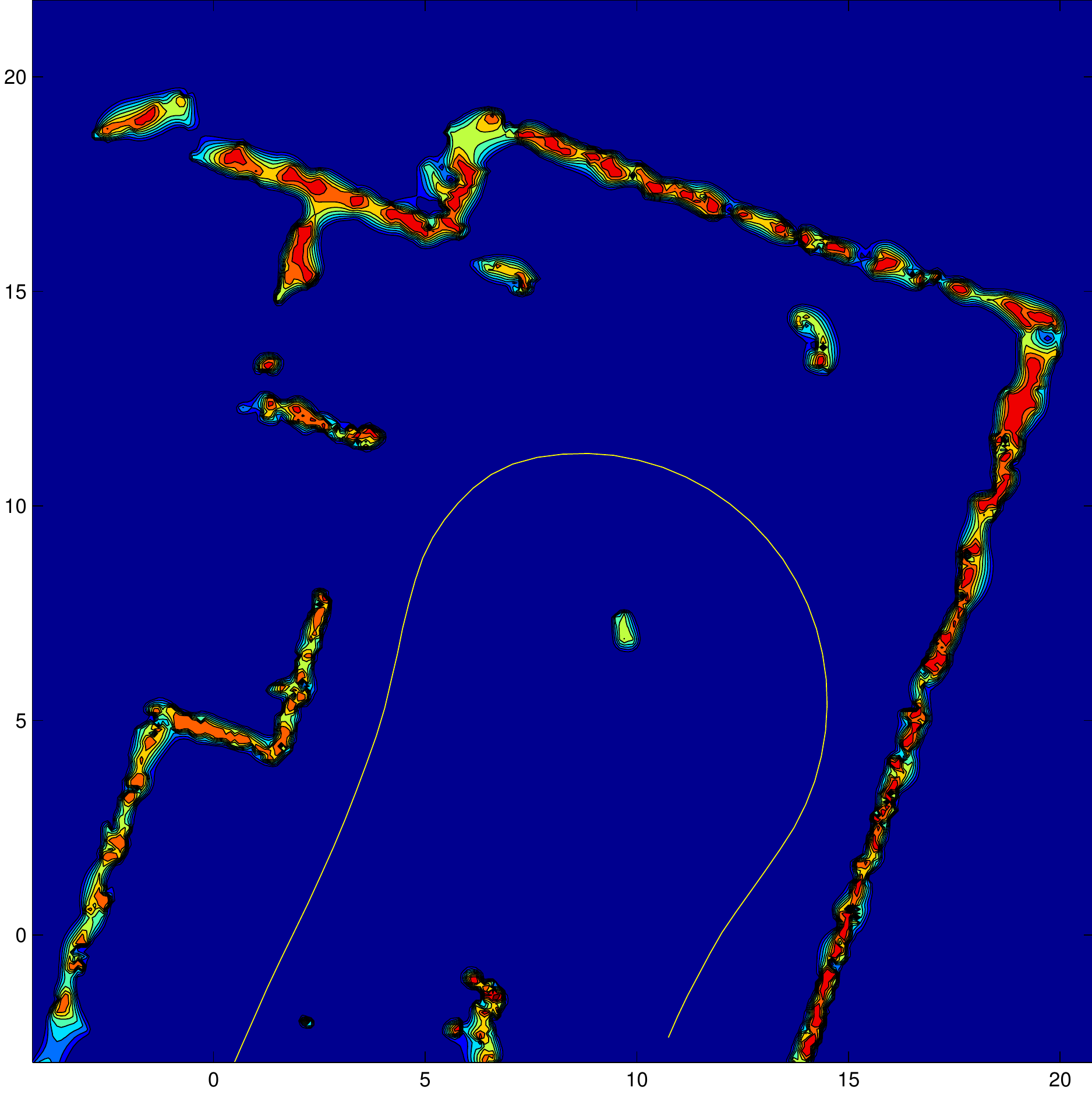}
		\includegraphics[width=2.25in]{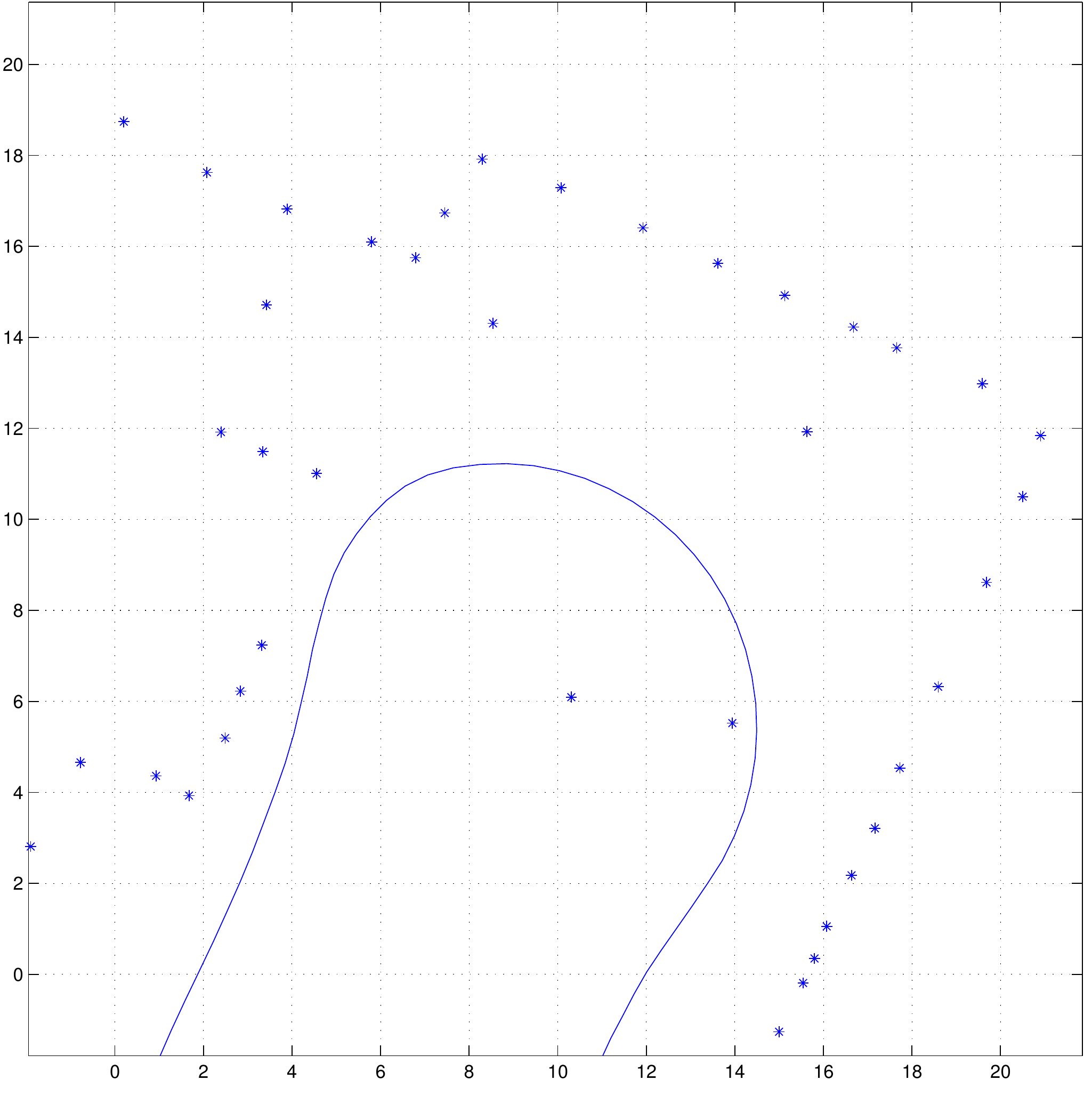}}
	\centerline{
		\includegraphics[width=2.25in]{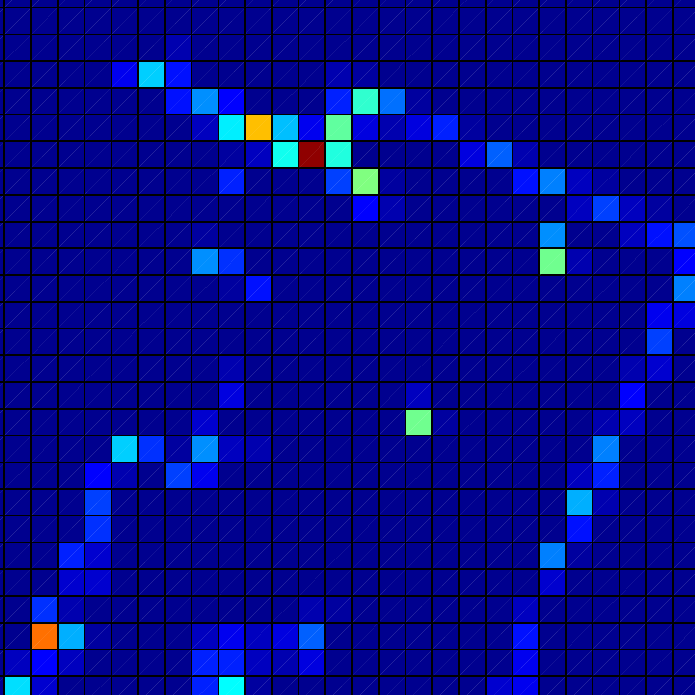}
		\includegraphics[width=2.25in]{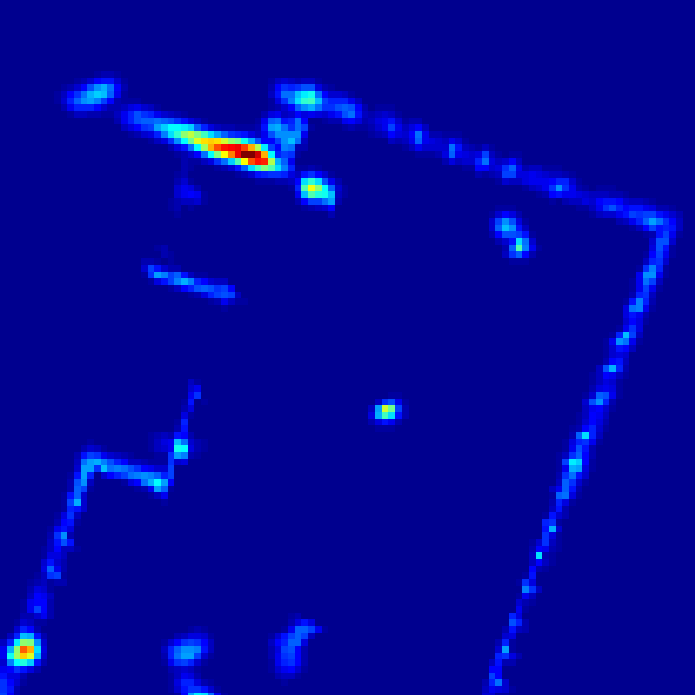}}
	\caption{A detailed contour graph of the map's probability distribution in contrast with the FastSLAM 2.0 and Grid based SLAM with low and high resolution. Colors blue to red corresponds to low to high probability. The yellow line represents the car's path estimation. a. G-SLAM with 3280 map points, b. FastSLAM 2.0, c. Grid SLAM with 3000 cells, d. High resolution Grid SLAM with 48000 cells.}
	\label{f2}
\end{figure}

\begin{figure}
	\centerline{\includegraphics[width=3.25in]{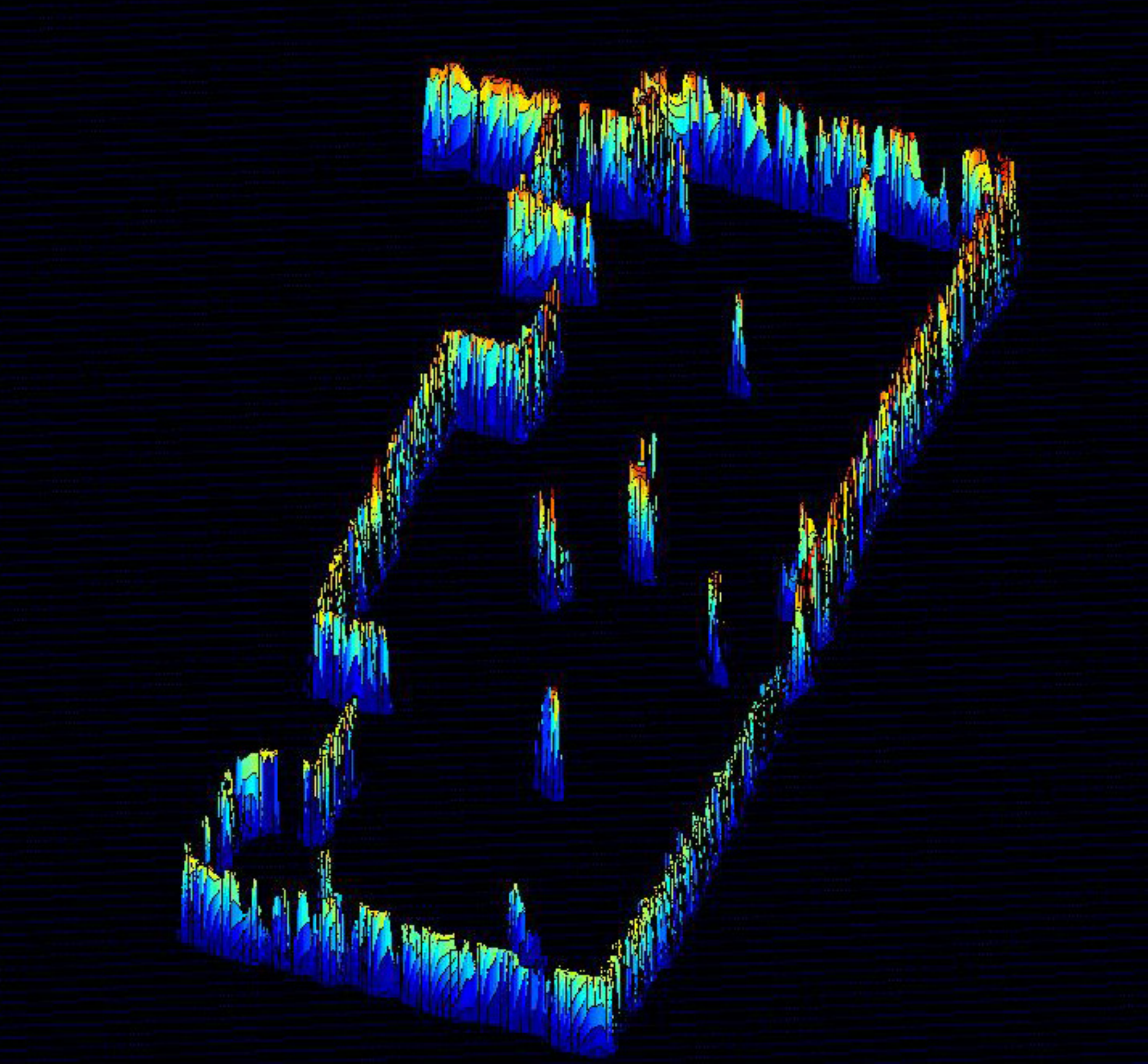}}
	\caption{Surface of the map's probability distribution as resulted from the G-SLAM method}
	\label{f3}
\end{figure}

Figures \ref{f1},\ref{f2},\ref{f3} demonstrates the map resulted from the G-SLAM method with $N=8$ particles and $M=10$ additional features for every observation, while the resulted map consisted of about 3280 map points (a mean density of $1.1$ map points per square meter).

Figure \ref{f2} shows a detailed view of the G-SLAM map probability distribution in comparison to the FastSLAM 2.0 map with 8 particles and the Grid SLAM with 3000 cells and a high resolution Grid SLAM with 48000 cells. 
It is noteworthy that the G-SLAM's map exhibits more detailed characteristics than the FastSLAM's and the low resolution Grid SLAM even if the Grid SLAM's cells are almost the same in number with the G-SLAM's map points. In order to achieve the same resolution with Grid SLAM we need to use around 15 times more dense grid with almost 48000 cells as it is shown in the fourth image on figure \ref{f2}. Table \ref{t4} shows that the G-SLAM method is slower and inaccurate with small amount of feature particles than FastSLAM, but on the other hand G-SLAM is more accurate and faster when is used with higher number of features $M$. Also the area which is free of obstacle (blue area) is also free of features $\theta$ and all of the features are gathered around the obstacles. The red area represents the highest possibility of the existence of obstacles. Figure \ref{f3} shows the surface of the map's probability distribution on the same run.

\begin{table}
	\caption{Comparison results between G-SLAM, FastSLAM 1.0 (FS 1), FastSLAM 2.0 (FS 2), Grid Occupancy SLAM (GOSLAM) and Grid Occupancy SLAM with high resolution (GOSLAM HR) on the Car Park dataset.}
	\begin{center}
		\begin{tabular}{c|c|c|c|c}
			\hline\hline

			Method		& Number of		& Number of	& Time/step	& Position \\
			& particles N 	& features M&(sec)		& error (m)\\ 
			\hline
			GSLAM 		&$2$ 			&  $4$		&  $22$ ms  &  $1.55$ m \\
			FS 1		&$2$ 			&  $-$  	&  $12$ ms  &  $0.84$ m \\
			FS 2		&$2$  			&  $-$  	&  $31$ ms  &  $0.47$ m \\
			GOSLAM		&$2$ 			&  $-$  	&  $14$ ms  &  $1.04$ m \\
			GOSLAM HR	&$2$  			&  $-$  	&  $180$ ms &  $1.10$ m \\
			\hline
			GSLAM 		&$8$ 			&  $10$  	&  $81$ ms	&  $0.41$ m \\
			FS 1		&$8$ 			&  $-$  	&  $51$ ms 	&  $0.62$ m \\
			FS 2		&$8$  			&  $-$  	&  $101$ ms	&  $0.42$ m \\
			GOSLAM		&$8$ 			&  $-$  	&  $46$ ms  &  $0.57$ m \\
			GOSLAM HR	&$8$  			&  $-$  	&  $648$ ms &  $0.41$ m \\
			\hline
			GSLAM 		&$30$ 			&  $10$  	&  $294$ ms	&  $0.40$ m \\
			FS 1		&$30$ 			&  $-$  	&  $148$ ms	&  $0.43$ m \\
			FS 2		&$30$  			&  $-$  	&  $281$ ms	&  $0.40$ m \\
			GOSLAM		&$30$ 			&  $-$  	&  $221$ ms &  $0.40$ m \\
			GOSLAM HR	&$30$  			&  $-$  	&  $1943$ ms&  $0.41$ m \\
			\hline\hline
		\end{tabular}
		\label{t4}
	\end{center}
\end{table}

Experiments performed with a variety in the number of particles $N$ and in the number of additional map points $M$. Table \ref{t3} presents the resulted mean distance error of the car and the mean process time using 2, 8 and 30 particles in the pose estimation procedure and 4, 10 and 20 additionally generated map points for every observation in the map update procedure.

\begin{table}
	\caption{Comparison results on the Car Park dataset with different number of particles and different number of per step additional points $M$.}
	\begin{center}
		\begin{tabular}{c|c|c|c}
			\hline\hline
			Number of	& Number of	& Time/step & Position \\
			particles N & features M&(ms)		& error (m)\\ 
			\hline
			
			$2$ 		&  $4$    	&  $22$ ms  &  $1.55$ m \\
			$2$ 		&  $10$  	&  $28$ ms  &  $1.15$ m \\
			$2$  		&  $20$  	&  $33$ ms  &  $1.19$ m \\
			\hline
			$8$ 		&  $4$  	&  $74$ ms &  $0.44$ m \\
			$8$ 		&  $10$  	&  $81$ ms &  $0.41$ m \\
			$8$  		&  $20$  	&  $92$ ms &  $0.40$ m \\
			\hline
			$30$ 		&  $4$  	&  $278$ ms &  $0.42$ m \\
			$30$ 		&  $10$  	&  $294$ ms &  $0.40$ m \\
			$30$  		&  $20$  	&  $314$ ms &  $0.40$ m \\
			\hline\hline
		\end{tabular}
		\label{t3}
	\end{center}
\end{table}

Table \ref{t3} shows that the G-SLAM algorithm results in a relatively high mean distance error when runs with 2 particles, due to its incapability to be consistent with few particles. In this case the map and the car's path acquire a cumulative error high enough to lead the algorithm into inconsistency. On the other hand the algorithm seems to converge relative fast with respect to the number of particles, since with 8 particle results in the minimum error.

\section{Conclusions}

In this paper it is presented the G-SLAM method for the estimation of the SLAM problem. This method is based on the simulation technique on both the kinematic and measurement models. Combining probabilities resulted from recursive forms the algorithm exports a detailed probability distribution of the map along with the best estimation of the robot's trajectory.

Future work will be the extension of the G-SLAM method in to dynamic environments. 
The method and techniques we have developed will be applied to a robotic platform and we will investigate the accuracy of the results and the consistency of the method in real case scenarios and dynamic environments.

\bibliographystyle{elsarticle-num}
\bibliography{GRAIN}

\begin{thebibliography}{10}
\expandafter\ifx\csname url\endcsname\relax
  \def\url#1{\texttt{#1}}\fi
\expandafter\ifx\csname urlprefix\endcsname\relax\def\urlprefix{URL }\fi
\expandafter\ifx\csname href\endcsname\relax
  \def\href#1#2{#2} \def\path#1{#1}\fi

\bibitem{57472}
R.~Smith, M.~Self, P.~Cheeseman, A stochastic map for uncertain spatial
  relationships, in: Proceedings of the 4th international symposium on Robotics
  Research, MIT Press, Cambridge, MA, USA, 1988, pp. 467--474.

\bibitem{Smith:1986fk}
R.~Smith, P.~Cheeseman, On the representation and estimation of spatial
  uncertainty.

\bibitem{Thrun:2005fk}
S.~Thrun, W.~Burgard, D.~Fox, Probabilistic Robotics (Intelligent Robotics and
  Autonomous Agents series), The MIT Press, 2005.

\bibitem{1241885}
M.~Montemerlo, S.~Thrun, Simultaneous localization and mapping with unknown
  data association using fastslam, in: Robotics and Automation, 2003.
  Proceedings. ICRA '03. IEEE International Conference on, Vol.~2, 2003, pp.
  1985--1991 vol.2.
\newblock \href {http://dx.doi.org/10.1109/ROBOT.2003.1241885}
  {\path{doi:10.1109/ROBOT.2003.1241885}}.

\bibitem{DBLP:conf/aaai/MontemerloTKW02}
M.~Montemerlo, S.~Thrun, D.~Koller, B.~Wegbreit, Fastslam: A factored solution
  to the simultaneous localization and mapping problem, in: AAAI/IAAI, 2002,
  pp. 593--598.

\bibitem{DBLP:conf/ijcai/MontemerloTKW03}
M.~Montemerlo, S.~Thrun, D.~Koller, B.~Wegbreit, Fastslam 2.0: An improved
  particle filtering algorithm for simultaneous localization and mapping that
  provably converges, in: IJCAI, 2003, pp. 1151--1156.

\bibitem{Montemerlo:2007kx}
M.~Montemerlo, S.~Thrun, B.~Siciliano, FastSLAM: a scalable method for the
  simultaneous localization and mapping problem in robotics, Vol. v. 27,
  Springer, Berlin, 2007.

\bibitem{Dissanayake:2001fk}
M.~Dissanayake, P.~Newman, S.~Clark, H.~Durrant-Whyte, M.~Csorba, A solution to
  the simultaneous localization and map building (slam) problem, Ieee
  Transactions On Robotics and Automation 17~(3) (2001) 229--241.

\bibitem{Grisetti:2007vn}
G.~Grisetti, C.~Stachniss, W.~Burgard, Improved techniques for grid mapping
  with rao-blackwellized particle filters, IEEE Transactions on Robotics 23~(1)
  (2007) 34--46.
\newblock \href {http://dx.doi.org/10.1109/TRO.2006.889486}
  {\path{doi:10.1109/TRO.2006.889486}}.

\bibitem{Rodriguez-Losada:fk}
D.~Rodriguez-Losada, P.~San~Segundo, F.~Matia, L.~Pedraza, Dual fastslam: Dual
  factorization of the particle filter based solution of the simultaneous
  localization and mapping problem, Journal of Intelligent and Robotic
  Systems\href {http://dx.doi.org/10.1007/s10846-008-9296-4}
  {\path{doi:10.1007/s10846-008-9296-4}}.

\bibitem{dpslam}
R.~P. Austin~Eliazar, Dp-slam: Fast, robust simultaneous localization and
  mapping without predetermined landmarks, in: International Joint Conference
  on Artificial Intelligence, 2003.

\bibitem{Zikos:2014aa}
N.~Zikos, V.~Petridis, \href{http://dx.doi.org/10.1007/s10846-014-0029-6}{6-dof
  low dimensionality slam (l-slam)}, Journal of Intelligent \& Robotic Systems
  (2014) 1--18.
\newline\urlprefix\url{http://dx.doi.org/10.1007/s10846-014-0029-6}

\bibitem{mywcci10}
V.~Petridis, N.~Zikos, L-slam: reduced dimensionality fastslam algorithms, in:
  WCCI, 2010, pp. 2510--2516.

\bibitem{6265692}
N.~Zikos, V.~Petridis, G-slam: A novel slam method, in: Control Automation
  (MED), 2012 20th Mediterranean Conference on, 2012, pp. 530--535.
\newblock \href {http://dx.doi.org/10.1109/MED.2012.6265692}
  {\path{doi:10.1109/MED.2012.6265692}}.

\bibitem{1348894}
N.~Kwak, G.-w. Kim, B.-h. Lee, A new compensation technique based on analysis
  of resampling process in fastslam, Robotica 26~(2) (2008) 205--217.
\newblock \href {http://dx.doi.org/http://dx.doi.org/10.1017/S0263574707003773}
  {\path{doi:http://dx.doi.org/10.1017/S0263574707003773}}.

\bibitem{Kehagias91convergenceproperties}
A.~Kehagias, Convergence properties of the lainiotis partition algorithm,
  Control and Computers 1 (1991) 6.

\bibitem{1241630}
J.~Nieto, J.~Guivant, E.~Nebot, S.~Thrun, Real time data association for
  fastslam, in: Robotics and Automation, 2003. Proceedings. ICRA '03. IEEE
  International Conference on, Vol.~1, 2003, pp. 412--418 vol.1.

\bibitem{Nieto04thehybrid}
J.~I. Nieto, J.~E. Guivant, E.~M. Nebot, The hybrid metric maps (hymms): A
  novel map representation for denseslam, in: In IEEE International Conference
  on Robotics and Automation (ICRA, 2004, pp. 391--396.

\end{thebibliography}

\end{document}